\documentclass[11pt,a4paper]{article}
\usepackage[hyperref]{acl2021}
\usepackage{times}
\usepackage{latexsym}

\usepackage{microtype}

\aclfinalcopy 
\def\aclpaperid{18} 

\usepackage{xspace}
\usepackage{subcaption}
\usepackage{latexsym}
\usepackage{amsmath}
\usepackage{multirow}
\usepackage{tablefootnote}
\usepackage{comment}
\usepackage{amssymb}
\usepackage{url}
\usepackage{enumitem}
\usepackage{array}
\usepackage{makecell}
\usepackage{adjustbox}
\usepackage{diagbox}
\setlist{nosep}

\usepackage{mdframed}

\usepackage{booktabs}
\usepackage{siunitx}
\renewrobustcmd{\bfseries}{\fontseries{b}\selectfont}
\robustify\bfseries

\usepackage{xcolor}

\newcommand{\Bert}{\textsc{Bert}\xspace}
\newcommand{\BertPGN}{\textsc{BertPgn}\xspace}
\usepackage{graphicx,calc}
\DeclareMathOperator*{\argmax}{arg\,max}
\DeclareMathOperator*{\LN}{LN}
\DeclareMathOperator*{\MHA}{MHA}
\DeclareMathOperator*{\FFN}{FFN}

\newcolumntype{C}[1]{>{\centering\arraybackslash}p{#1}}

\title{Summary-Oriented Question Generation for Informational Queries}

\author{Xusen Yin\thanks{~~Work was done as an intern at Amazon.} \\
  USC/ISI \\
  \texttt{xusenyin@isi.edu} \\\And
  Li Zhou \\
  Amazon \\
  \texttt{lizhouml@amazon.com} \\\AND
  Kevin Small \\
  Amazon \\
  \texttt{smakevin@amazon.com} \\\And
  Jonathan May \\
  USC/ISI \\
  \texttt{jonmay@isi.edu} \\
  }

\date{}

\begin{document}
\maketitle
\begin{abstract}
Users frequently ask simple factoid questions for question answering (QA) systems, attenuating the impact of myriad recent works that support more complex questions. Prompting users with automatically generated {\em suggested questions} (SQs) can improve user understanding of QA system capabilities and thus facilitate more effective use. We aim to produce self-explanatory questions that focus on main document topics and are answerable with variable length passages as appropriate. We satisfy these requirements by using a \Bert-based Pointer-Generator Network trained on the Natural Questions (NQ) dataset. Our model shows SOTA performance of SQ generation on the NQ dataset (\num{20.1} BLEU-4).
We further apply our model on out-of-domain news articles, evaluating with a QA system due to the lack of gold questions and demonstrate that our model produces better SQs for news articles -- with further confirmation via a human evaluation.
\end{abstract}

\section{Introduction}

Question answering (QA) systems have experienced dramatic recent empirical improvements due to several factors including novel neural architectures~\cite{chen2020open}, access to pre-trained contextualized embeddings~\cite{devlin-etal-2019-bert}, and the development of large QA training corpora~\cite{rajpurkar-etal-2016-squad,trischler-etal-2017-newsqa,10.1145/3366423.3380114}. However, despite technological advancements that support more sophisticated questions~\cite{yang-etal-2018-hotpotqa,joshi-etal-2017-triviaqa,choi-etal-2018-quac,reddy-etal-2019-coqa}, many consumers of QA technology in practice tend to ask simple factoid questions when engaging with these systems. Potential explanations for this phenomenon include low expectations set by previous QA systems, limited coverage for more complex questions not changing these expectations, and users simply not possessing sufficient knowledge of the subject of interest to ask more challenging questions. Irrespective of the reason, one potential solution to this dilemma is to provide users with automatically generated {\em suggested questions} (SQs) to help users better understand QA system capabilities.

Generating SQs is a specific form of question generation (QG), a long-studied task with many applied use cases -- the most frequent purpose being data augmentation for mitigating the high sample complexity of neural QA models~\cite{alberti-etal-2019-synthetic}. 
However, the objective of such existing QG systems is to produce large quantities of question/answer pairs for training, which is contrary to that of SQs. The latter seeks to guide users in their research of a particular subject by producing engaging and understandable questions. To this end, we aim to generate questions that are {\em self-explanatory} and {\em introductory}.

{\em Self-explanatory questions} require neither significant background knowledge nor access to documents used for QG to understand the SQ context. For example, existing QG systems may use the text {\em ``On December 13, 2013, Beyonc\'{e} unexpectedly released her eponymous fifth studio album on the iTunes store without any prior announcement or promotion."} to produce the question {\em ``Where was the album released?"} This kind of question is not uncommon in crowd-sourced datasets (e.g., SQuAD~\cite{rajpurkar-etal-2016-squad}) but do not satisfy the self-explanatory requirement. \newcite{clark-gardner-2018-simple} estimate that \SI{33}{\percent} of SQuAD questions are context-dependent. This context-dependency is not surprising, given that annotators observe the underlying documents when generating questions. 

{\em Introductory questions} are best answered by a larger passage than short spans such that users can learn more about the subject, possibly inspiring follow-up questions (e.g., ``Can convalescent plasma help COVID patients?"). 
However, existing QG methods mostly generate questions while reading the text corpus and tend to produce narrowly focused questions with close syntactic relations to associated answer spans.
TriviaQA \cite{joshi-etal-2017-triviaqa} and HotpotQA \cite{yang-etal-2018-hotpotqa} also provide fine-grained questions, even though reasoning from a larger document context via multi-hop inference. This narrower focus often produces factoid questions peripheral to the main topic of the underlying document and is less useful to a human user seeking information about a target concept.

Conversely, the Natural Question (NQ) dataset \cite{kwiatkowshi-etal-2019-natural-questions} (and similar ones such as MS Marco \cite{bajaj2016ms}, GooAQ~\cite{khashabi2021gooaq}) is significantly closer to simulating the desired information-seeking behavior. Questions are generated independently of the corpus by processing search query logs, and the resulting answers can be entities, spans in texts (aka short answers), or entire paragraphs (aka long answers). Thus, the NQ dataset is more suitable as QG training data for generating SQs as long-answer questions that tend to satisfy our self-explanatory and introductory requirements.

To this end, we propose a novel \Bert-based Pointer-Generator Network (\BertPGN) trained with the NQ dataset to generate introductory and self-explanatory questions as SQs. Using NQ, we start by creating a QG dataset that contains questions with both short and long answers.
We train our \BertPGN model with these two types of context-question pairs together. During inference, the model can generate either short- or long-answer questions as determined by the context.
With automatic evaluation metrics such as BLEU~\cite{10.3115/1073083.1073135}, we show that for long-answer question generation, our model can produce state-of-the-art performance with \num{20.1} BLEU-4, \num{6.2} higher than \cite{mishra2020automatic}, the current state-of-the-art on this dataset. The short answer question generation performance can reach \num{28.1} BLEU-4.

We further validate the generalization ability of our \BertPGN model by creating an out-of-domain test set with the CNN/Daily Mail \cite{10.5555/2969239.2969428}.
Without human-generated reference questions, automatic evaluation metrics such as BLEU are not usable.
We propose to evaluate these questions with a pretrained QA system that produces two novel metrics. The first is {\it answerability}, measuring the possibility to find answers from given contexts. The second is {\it granularity}, indicating whether the answer would be passages or short spans.
Finally, we conduct a human evaluation with generated questions of the test set and demonstrate that our \BertPGN~model can produce introductory and self-explanatory questions for information-seeking scenarios, even for a new domain that differs from the training data.

The novel contributions of our paper include:

\begin{itemize}
    \item We generate questions, aiming to be both introductory and self-explanatory, to support human information seeking QA sessions.
    \item We propose to use the \Bert-based Pointer-Generator Network to generate questions by encoding larger contexts capable of resulting in answer forms including entities, short text spans, and even whole paragraphs.
    \item We evaluate our method, both automatically and with human evaluation, on in-domain Natural Questions and out-of-domain news datasets, providing insights into question generation for information seeking.
    \item We propose a novel evaluation metric with a pretrained QA system for generated SQs when there is no reference question.
\end{itemize}

\section{Related Work}

QG has been studied in multiple application contexts (e.g., generating questions for reading comprehension tests~\cite{heilman-2010-good}, generating questions about an image~\cite{mostafazadeh-etal-2016-generating}, recommending questions with respect to a news article~\cite{laban-etal-2020-whats}), evaluating summaries \cite{deutsch2020questionanswering,wang-etal-2020-asking}, and using multiple methods~(see \cite{pan-etal-2019} for a recent survey). Early neural models focused on sequence-to-sequence generation based solutions~\cite{serban-etal-2016-generating,du-etal-2017-learning}. The primary directions for improving these early works generally fall into the categories of providing mechanisms to inject answer-aware information into the neural encoder-decoder architectures~\cite{du-cardie-2018-harvesting,li-etal-2019-improving-question,10.1145/3308558.3313737,wang-etal-2020-asking,sun-etal-2018-answer}, encoding larger portions of the answer document as context~\cite{zhao-etal-2018-paragraph,Tuan_2020}, and incorporating richer knowledge sources~\cite{elsahar-etal-2018-zero}.

These QG methods and the work described in this paper focus on using single-hop QA datasets such as SQuAD~\cite{rajpurkar-etal-2016-squad,rajpurkar-etal-2018-know}, NewsQA~\cite{trischler-etal-2017-newsqa,10.5555/2969239.2969428}, and MS Marco~\cite{bajaj2016ms}. However, there has also been recent interest in multi-hop QG settings~\cite{10.1145/3366423.3380114,gupta2020reinforced,malon2020generating} by using multi-hop QA datasets including HotPotQA~\cite{yang-etal-2018-hotpotqa}, TriviaQA~\cite{joshi-etal-2017-triviaqa}, and FreebaseQA~\cite{jiang-etal-2019-freebaseqa}. Finally, there has been some recent interesting work regarding {\em unsupervised} QG, where the goal is to generate QA training data without an existing QG corpus to train better QA models~\cite{lewis-etal-2019-unsupervised,li2020harvesting}.

Most directly related to our work from a motivation perspective is recent research regarding providing SQs in the context of supporting a news chatbot~\cite{laban-etal-2020-whats}. However, the focus of this work is not QG, where they essentially use a GPT-2 language model~\cite{radford2019language} trained on SQuAD data for QG and do not evaluate this component independently. 
\newcite{qi2020stay} generates questions for information-seeking but not focuses on introductory questions.
Most directly related to our work from a conceptual perspective is regarding producing questions for long answer targets~\cite{mishra2020automatic}, which we contrast directly in Section~\ref{sec:definition}.  As QG is a generation task, automated evaluation frequently uses metrics such as BLEU \cite{10.3115/1073083.1073135}, METEOR \cite{10.5555/1626355.1626389}, and ROUGE \cite{lin-2004-rouge}. As these do not explicitly evaluate the requirements of our information-seeking use case, we also evaluate using the output of a trained QA system and conduct human annotator evaluations. 

\section{Problem Definition}
\label{sec:definition}

Given a context $X$ and an answer $A$, we want to generate a question $\tilde{Q}$ that satisfies
\[\tilde{Q} = \argmax_{Q} P(Q | X, A),\]
where the context $X$ could be a paragraph or a document that contains answers, rather than sentences as used in \cite{du-cardie-2018-harvesting,Tuan_2020}, while $A$ could be either short spans in $X$ such as entities or noun phrases (referred to as a {\it short answer}), or the entire context $X$ (referred to as a {\it long answer}).

The {\it long answer} QG task targets generating questions that are best answered by the entire context (i.e., paragraph or document) or a summary of the context, which is notably different from most QG settings where the answer is a short text span and the context is frequently a single sentence.
\newcite{mishra2020automatic} also work on the {\it long answer} QG setting using the NQ dataset, but their task definition is $\argmax_{Q} P(Q | X)$ where they refer to the context $X$ as the {\it long answer}. We use their models as baselines. 


\section{Methods}
\begin{figure}[t]
    \centering
    \small
    \includegraphics[width=0.9\columnwidth]{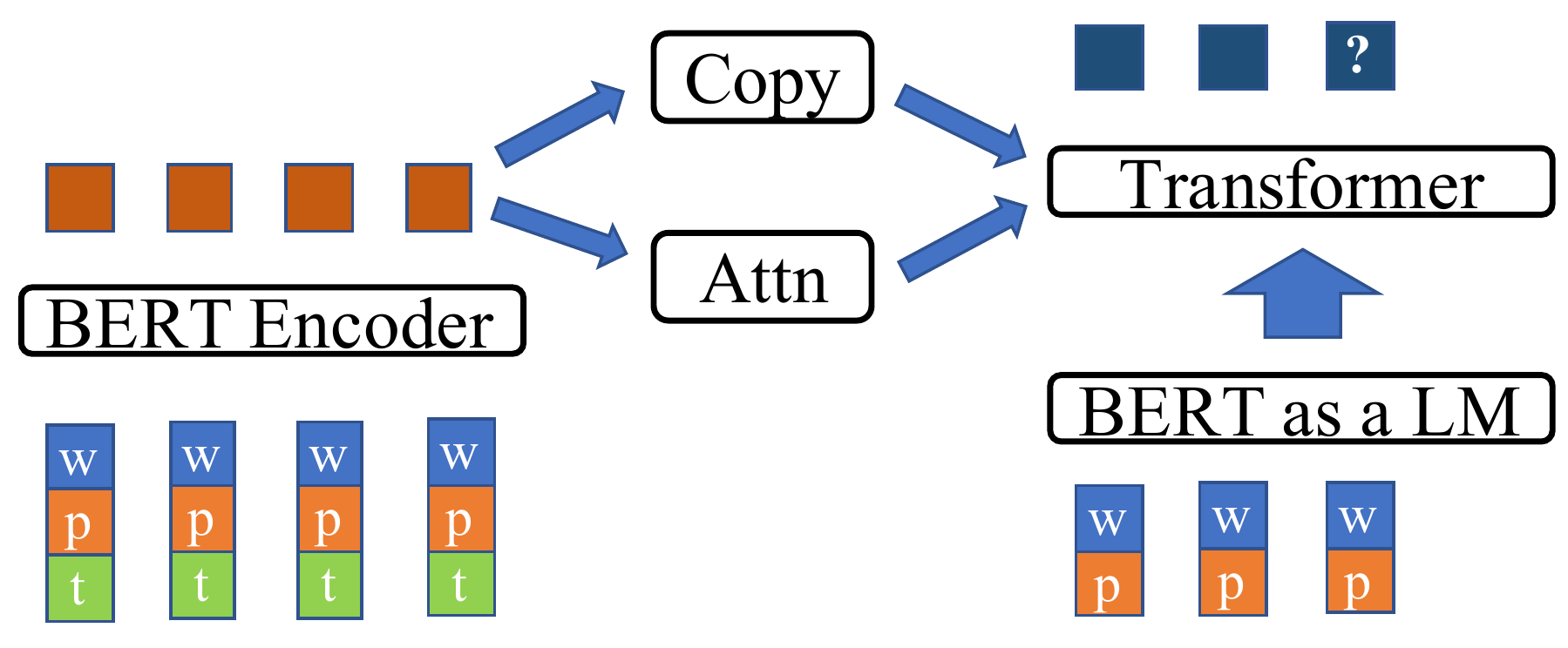}
    \caption{The \BertPGN~architecture. The input for the \Bert~encoder is the context (w/p: word and position embeddinngs) with answer spans (or the whole context in the long answer setting) marked with the answer tagging (t: answer tagging embeddings). The decoder is a combination of \Bert~as a language model (i.e. has only self-attentions) and a Transformer-based pointer-generator network.}
    \label{fig:bertpgn-architecture}
\end{figure}

We use the \Bert-based Pointer-Generator Network (\BertPGN) to generate questions. \newcite{Tuan_2020} use two-layer cross attentions between contexts and answers to encode contexts such as paragraphs when generating questions and show improved results. However, they show that three-layer cross attentions produce worse results. We will show later in the experiment that this is due to a lack of better initialization and that a higher layer is better for long answer question generation. \newcite{zhao-etal-2018-paragraph} use answer tagging from the context instead of combining context and answer. Our model is motivated by these two works (Figure \ref{fig:bertpgn-architecture}).

\subsection{Context and Answer Encoding}

Given context $X=\{x_i\}_{i=1}^{L}$, we add positional embeddings $P=\{p_i\}_{i=1}^L$ and type embeddings $T=\{t_i\}_{i=1}^L$ as the input for \Bert. We use type embeddings to discriminate between a context and an answer, following \newcite{zhao-etal-2018-paragraph,Tuan_2020}. 
We use $t_i=0$ to represent `{\em context-only}' and $t_i=1$ to represent `{\em both context and answer}' for token $x_i$. 
We do not apply the \texttt{[CLS]} in the beginning since we do not need the pooled output from \Bert. We do not use the \texttt{[SEP]} to combine contexts and answers as inputs for \Bert since we mark answers in the context with type embeddings. 
The sequence output from \Bert~which forms our context-answer encoding is given by
\[H = f_{\Bert} \left(X+P+T\right).\]

\subsection{Question Decoding}

The transformer-based Pointer-Generator Network is derived from \cite{see-etal-2017-get} with adaptations to support transformers \cite{NIPS2017_3f5ee243}.
Denoting $\LN(\cdot)$ as layer normalization, $\MHA(Q,K,V)$ as the multi-head attention with three parameters---query, key, and value, $\FFN(\cdot)$ as a linear function, and the decoder input at time $t$: $Y^{(t)}=\{y_j\}_{j=1}^t$, the decoder self-attention at time $t$ is given by (illustrated with a single-layer transformer simplification)
\[A_S^{(t)} = \LN\left(\MHA\left(Y^{(t)}, Y^{(t)}, Y^{(t)}\right) + Y^{(t)}\right),\]
the cross-attention between encoder and decoder is
\[A_C^{(t)} = \LN\left(\MHA\left(A_S^{(t)}, H, H\right) + A_S^{(t)}\right),\]
and the final decoder output is
\[O^{(t)} = \LN\left(\FFN\left(A_C^{(t)}\right) + A_C^{(t)}\right).\]

Using the LSTM \cite{lstm-original} encoder-decoder model, \newcite{see-etal-2017-get} compute a generation probability using the encoder context, decoder state, and decoder input. While the transformer decoder cross-attention $A_C^{(t)}$ already contains a linear combination between self-attention of decoder input and encoder-decoder cross attention. Thus, we use the combination of the decoder input and cross-attention to compute the generation probability
\[P_G^{(t)} = \FFN\left(\left[Y^{(t)}, A_C^{(t)}\right]\right).\]

To improve generalization, we also use a separate \Bert~model as a language model (LM) for the decoder. Even though \Bert~is not trained to predict the next token \cite{devlin-etal-2019-bert} as with typical language models (e.g., GPT-2), we still choose \Bert~as our LM to ensure the COPY mechanism shares the same vocabulary between the encoder and the decoder.\footnote{Note that we change the masking for the original \Bert when using \Bert as a LM, since the decoder at step $t$ should not read inputs at steps $t+i$ where $i\ge 0$.} We also do not need to process out-of-vocabulary words because we use the BPE \cite{sennrich-etal-2016-neural,devlin-etal-2019-bert} tokenization in both the encoder and decoder.

\section{Dataset}

\subsection{Natural Questions dataset}
We use Natural Questions dataset \cite{kwiatkowshi-etal-2019-natural-questions} for training as NQ questions are independent of their supporting documents.
NQ has \num{307000} training examples, answered and annotated from Wikipedia pages, in a format of a question, a Wikipedia link, long answer candidates, and short answer annotations. 
\SI{51}{\percent} of these questions have no answer for either being invalid or non-evidence in their supporting documents.
Another \SI{36}{\percent} have long answers that are paragraphs and have corresponding short answers that either spans long answers or being masked as yes-or-no.
The remaining \SI{13}{\percent} questions only have long answers. 
We are most interested in the last portion of questions as they are best answered by summaries of their long answers, reflecting the coarse-grained information-seeking behavior.\footnote{Data annotation is a {\it subjective} task where different annotators could have different opinions for whether there is a short answer or not. NQ uses multi-fold annotations (e.g., a \num{5}-fold annotation for the dev set). However, the training data only has the \num{1}-fold annotation, so whether there is a short answer is not \SI{100}{\percent} accurate.}

We use paragraphs that contain long answers or short answers in NQ as the context. 
We do not consider using the whole Wikipedia page, i.e., the document, as the context as most Wikipedia pages are too long to encode: In the NQ training set, there are \num{8407} tokens at document level on average, while for news articles in the CNN/Daily Mail that we will discuss in Section \ref{sec:newsqa-data}, the average document size is \num{583} \cite{Tuan_2020}, which is not much larger than the average size of long answers in NQ (\num{384} tokens).

We also consider the ratio between questions and the context-answer pairs to avoid generating multiple questions based on the same context-answer. After removing questions that have no answers, there are \num{152148} questions and \num{136450} unique long answers. The average ratio between questions and long answers is around \num{1.1} questions per {\it paragraph} (ratios are in a range of \numrange{1}{47}). The average ratio is more reasonable for question generation, comparing to the SQuAD where there are \num{1.4} questions per {\it sentence} on average \cite{du-etal-2017-learning}.

\subsubsection{NQ Preprocessing}

\begin{table}[t]
\centering
\small
\begin{tabular}{llS[table-format=8]}
\toprule
data     & type           & {count} \\
\midrule
train    & mix            & 99725 \\
dev      & mix            & 11140 \\
NQ-SA  & long and short & 3364  \\
NQ-LA  & long only      & 1495  \\
News-LA & long only & 3048\\
\bottomrule
\end{tabular}
\caption{QG Data summary. *-LA contains questions that only have long answers, while NQ-SA contains questions having both long and short answers.}
\label{tbl:data-summary}
\end{table}

We extract questions, long answers, and short answer spans from the NQ dataset. 
We also extract the Wikipedia titles since long answers alone do not always contain the words from their corresponding titles.
We add brackets (`\texttt{[}' and `\texttt{]}') for all possible short answer spans such that we can later extract these spans accordingly to avoid potential position changes due to context preprocessing (e.g.,  different tokenization).\footnote{Using brackets here is an arbitrary but functional choice.}
When there is no short answer, we add brackets to the whole long answer.
We then concatenate the titles with long answers as contexts.
For details, see examples from Figure \ref{fig:example-of-preprocessing} and Figure \ref{fig:bert-pgn-example} in Appendix \ref{sec:appendix}.

As in \cite{mishra2020automatic}, we only keep questions with long answers starting from the HTML paragraph tag. After preprocessing (Table \ref{tbl:data-summary}), we get \num{110865} question-context pairs, while \newcite{mishra2020automatic} gets \num{77501} pairs since they only keep long answer questions. We split the dataset with a \num{90}/\num{10} ratio for training/validation.

We use the original NQ dev set, which contains \num{7830} questions, as our test set. We follow the same extraction procedure as with the training and validation data modulo two new steps. First, noting that \SI{79}{\percent} of Wikipedia pages appearing in the NQ dev set are also present in the NQ training set, we filter all overlapped contexts from the NQ dev set when creating our test set. Second, the original NQ dev set is \num{5}-way annotated; thus, each question may have up to five different long/short answers. We treat each annotation as an independent context, even though they are associated with the same target question. To separately evaluate the QG performance for long answers and short answers, we split test data into {\it long-answer} questions (NQ-LA) and {\it short-answer} questions (NQ-SA). Finally, we get \num{4859} test data in total, with \num{1495} of them only have long answers while the remaining \num{3364} have both long and short answers while \newcite{mishra2020automatic} gets \num{2136} test data from the original dev set.

\subsection{News dataset}
\label{sec:newsqa-data}

We use the \num{12744} CNN news articles from the CNN/Daily Mail dataset~\cite{10.5555/2969239.2969428}) for the out-of-domain evaluation. 
We apply the same preprocessing method as in the NQ dataset to create a long-answer test set --- News-LA. We use whole news articles, instead of paragraphs, as contexts, considering to generate questions that lead to entire news articles as answers. For each news article, we first remove {\it highlights}, which is a human-generated summary, and datelines (e.g., NEW DELHI, India (CNN)). We filter out those news articles that are longer than \num{490} tokens with the BEP tokenization and those overlapped context-question pairs.
Finally, we get \num{3048} data in the News-LA test set.

\section{In-Domain Evaluation with Generation Metrics}

\subsection{Experiment Setup and Training}
\label{sec:train-bertpgn}

We use a \Bert-base uncased model \cite{devlin-etal-2019-bert} that contains \num{12} hidden layers. The vocabulary contains \num{30522} tokens. We create the PGN decoder with another \Bert~model from the same setting, followed by a \num{2}-layer transformer with \num{12} heads and \num{3072} intermediate sizes. The maximum allowed context length is \num{500}, while the maximum question length is \num{50}. We train our model on an Amazon EC2 P3 machine with one Tesla V100 GPU, with the batch size \num{10}, and the learning rate \num{5e-5} with the Adam optimizer \cite{DBLP:journals/corr/KingmaB14} on all parameters of the \BertPGN~model (both \Bert~models are trainable).
We train \num{20} epochs of our model and evaluate with the dev set to select the model according to perplexity. Each epoch takes around \num{20} minutes to finish.
Throughout the paper, we use the implementation of BLEU, METEOR, and ROUGE\_L by \newcite{sharma2017nlgeval}.

\subsection{In-Domain Evaluation}

We first evaluate our model using BLEU, METEOR, and ROUGE\_L to compare with \newcite{mishra2020automatic} on long answers (first two rows in Table \ref{tbl:bleu-compare}). 
The transformer-based iwslt\_de\_en is a German to English translation model with \num{6} encoder and decoder layers, \num{16} encoder and decoder attention heads, \num{1024} embedding dimension, and \num{4096} embedding dimension of feed forward network. 
The other transformer-based multi-source method, which is based on \cite{libovicky-etal-2018-input}, combines each context with a retrieval-based summary as input.
We decode questions from our model using beam search (beam=\num{3}).\footnote{ \newcite{mishra2020automatic} have not described the decoding method and possible beam size, but they use models from \cite{ott-etal-2018-scaling} that uses beam=\num{4}.} Evaluating on NQ-LA, our \BertPGN~model outperforms both existing models substantially with near seven points for all metrics. The performance for short answer questions NQ-SA is even better, with near eight more BLEU-4 points than NQ-LA.

\begin{table}[t]
\centering
\small
\begin{tabular}{lS[table-format=2.1]S[table-format=2.1]S[table-format=2.1]S[table-format=2.1]}
\toprule
     & {B1} & {B4} & {ME} & {RL} \\
\midrule
\makecell[l]{TX iwslt\_de\_en} & 36.8  & 13.9  & 17.5  & 35.6    \\
\makecell[l]{TX Multi-Source} & 36.0  & 13.3  & 16.8  & 34.6    \\
\makecell[l]{\BertPGN~LA} & \bfseries 43.9 & \bfseries 20.1 & \bfseries 22.6 & \bfseries 42.2  \\
\makecell[l]{\BertPGN~SA} & \bfseries 54.7 & \bfseries 28.1 & \bfseries 27.9 & \bfseries 53.2  \\
\bottomrule
\end{tabular}
\caption{Comparing our model (\BertPGN) on NQ-LA and NQ-SA with two models in \cite{mishra2020automatic}---their best performing Transformer\_iwslt\_de\_en and multi-source transformer combining contexts and automatically generated summaries, with automatic evaluation BLEU-1, BLEU-4, METEOR, and Rouge\_L.}
\label{tbl:bleu-compare}
\end{table}

\subsection{Ablation Study}

\begin{table}[t]
\centering
\small
\begin{tabular}{lS[table-format=2.1]S[table-format=2.1]}
\toprule
  B4   & {NQ-LA} & {NQ-SA} \\
\midrule
no-pointer & 17.1 & 23.6 \\
no-\Bert-LM (*) & 18.9 & 26.5 \\
* - no-type-id & 19.0 & 20.8 \\
* - no-init & 15.3 & 19.3 \\
* - 2-layer & 14.9 & 19.1 \\
\bottomrule
\end{tabular}
\caption{Ablation study of the \BertPGN. Removing the pointer network drops BLEU-4 by around \num{3} points for both test sets. Removing \Bert~initialization affects both the NQ-LA and NQ-SA substantially but more mildly than removing the pointer. Removing type IDs affects the NQ-SA by \num{5.7} drop in BLEU-4.}
\label{tbl:ablation-bert-pgn}
\end{table}

We first examine the effect of the pointer network from the \BertPGN. 
We then run ablation study by first removing \Bert-LM in the decoder, and independently
\begin{itemize}
    \item removing type IDs from \Bert~encoder
    \item removing \Bert~initialization for \Bert~encoder
    \item substituting \Bert~encoder with a 2-layer transformer
\end{itemize}

We train our \BertPGN~models from scratch for each setting and conduct these ablation studies for NQ-LA and NQ-SA separately (Table \ref{tbl:ablation-bert-pgn}).

Removing the pointer from the \BertPGN makes the BLEU-4 scores drop for both NQ-LA and NQ-SA more than removing the \Bert~as the LM in the decoder.
Type IDs are more helpful for NQ-SA (approximately a \num{5}-point drop in BLEU-4) than NQ-LA since NQ-SA needs to use type IDs to mark answers. Removing \Bert~initialization causes notable drops for both NQ-LA (\num{3.6} drops in BLEU-4) and NQ-SA (\num{7.2} in BLEU-4), which implies that \Bert achieves better generalization when encoding these considerably long contexts. Another interesting finding is that the NQ-LA is more sensitive to the number of layers of the encoder than NQ-SA. When decreasing the layers to two from twelve, NQ-LA drops by \num{0.4} in BLEU-4 while NQ-SA drops by \num{0.2}.

\section{Out-of-Domain Evaluation with QA Systems}
\label{sec:newsqa-la-evaluation}

We use a well-trained question answering system as the evaluation method, given that the automated scoring metrics have two notable drawbacks when evaluating long-answer questions: 
(1) There are usually multiple valid questions for long-answer question generation as contexts are much longer than previous work. However, most datasets only have one gold question for each context;
(2) They cannot measure generated questions when there is no gold question, which is the right problem that we encountered for our News-LA dataset.

\begin{figure}[t]
    \centering
    \small
    \includegraphics[width=0.65\columnwidth]{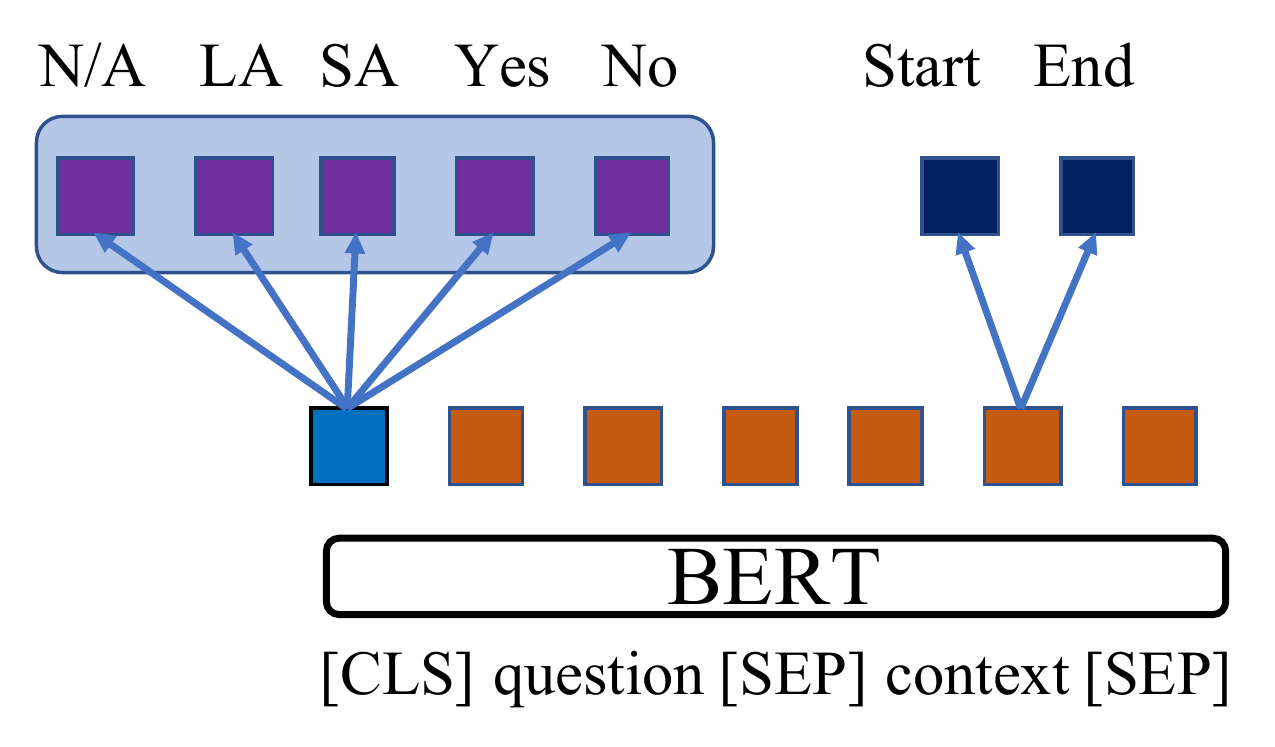}
    \caption{The \Bert-joint architecture \cite{alberti2019bert}. Input is the combined question and context, and the outputs are an answer-type classification from the \texttt{[CLS]} token and start/end of answer spans for each token from the context.}
    \label{fig:bert-joint-architecture}
\end{figure}

\subsection{The QA Metrics}

We use the \Bert-joint model \cite{alberti2019bert} (Figure \ref{fig:bert-joint-architecture}) for NQ question answering to evaluate our long answer question generation.
The \Bert-joint model takes the combination a question and the corresponding context as an input, outputs the probability of answer spans and the probability of answer types. 
For a context of size $n$, it produces $p_{start}$ and $p_{end}$ for each token, indicating whether this token is a start or end token of an answer span. 
It then chooses the answer span $(i, j)$ where $i < j$ that maximizes $p_{start}(i) \cdot p_{end}(j)$ as the probability of the answer. 
It also defines the probability of no answer to be $p_{start}([CLS])\cdot p_{end}([CLS])$, i.e., an answer span that starts then stops at the \texttt{[CLS]} token. 
Furthermore, the \Bert-joint model computes the probability of \textit{types} of the question---\textit{undetermined}, \textit{long answer}, \textit{short answer}, and \textit{YES-or-NO}. 
This model achieves \SI{66.2}{\percent} F1 on NQ long answer test set, which is \SI{10}{\percent} better compared to models used in \cite{kwiatkowshi-etal-2019-natural-questions,parikh-etal-2016-decomposable}. 
We define the {\it answerability} score ($s_{ans}$) as $\log \left(p_{ans} / p_{no\_ans}\right)$, and the {\it granularity} score ($s_{gra}$) as $\log \left(p_{la} / p_{sa}\right)$ when evaluating our long answer question generation with the \Bert-joint model.

\subsection{QG Models to Compare}
We construct a baseline model to compare as follows.
Using the same \BertPGN~architecture, we train a model on the SQuAD sentence-question pairs prepared by \newcite{du-etal-2017-learning}. When generating questions for news articles, we use the first line of each news article as the context, with the assumption that the first line is a genuine summary produced by humans.
Notice that the resulting baseline is the state-of-the-art for answer-free (the model does not know the whereabouts of answer spans) question generation with SQuAD (Table \ref{tbl:baseline}).
We refer to the model as $M_{SD}$ hereafter.
Similarly, we call our \BertPGN model trained on the NQ dataset as $M_{NQ}$.
We use beam search (beam=\num{3}) for both models.

\begin{table}[t]
\centering
\small
\begin{tabular}{lS[table-format=2.1]S[table-format=2.1]S[table-format=2.1]S[table-format=2.1]}
\toprule
           & {B1}             & {B4}             & {ME}             & {RL}             \\
\midrule
Du-17 best & 43.1          & 12.3          & 16.6          & 39.8          \\
 $M_{SD}$  & \bfseries 46.0 & \bfseries 14.8 & \bfseries 19.2 & \bfseries 42.0 \\
\bottomrule
\end{tabular}
\caption{The performance of our answer-free baseline, compared with the best model from \cite{du-etal-2017-learning}.}
\label{tbl:baseline}
\end{table}

\subsection{Evaluation Results}

\begin{figure}[t]
    \centering
    \small
    \includegraphics[trim=10mm 3mm 3mm 5mm,width=0.45\textwidth]{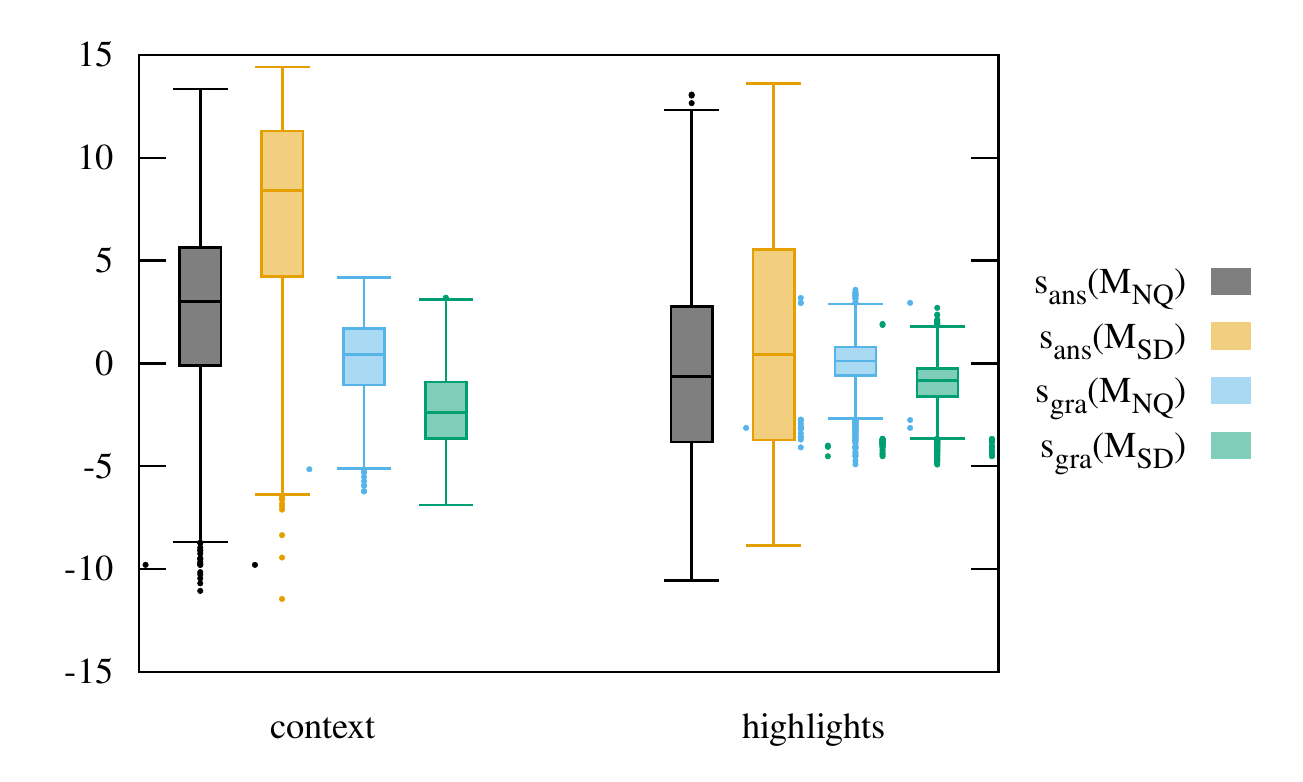}
    \caption{Answerability and granularity scores of generated questions for News-LA with the \Bert-joint model \cite{alberti2019bert} as the evaluation QA model by answering generated questions from either news article {\it context} or news article {\it highlights}. We compare two models: (1) NQ: \BertPGN~trained with NQ dataset and generate on whole news articles; (2) SD: \BertPGN~trained with SQuAD dataset and generate on the first line of each news article.}
    \label{fig:newsqa-scores}
\end{figure}

\begin{figure}
\centering
\small
   \includegraphics[trim=10mm 3mm 3mm 5mm,width=0.45\textwidth]{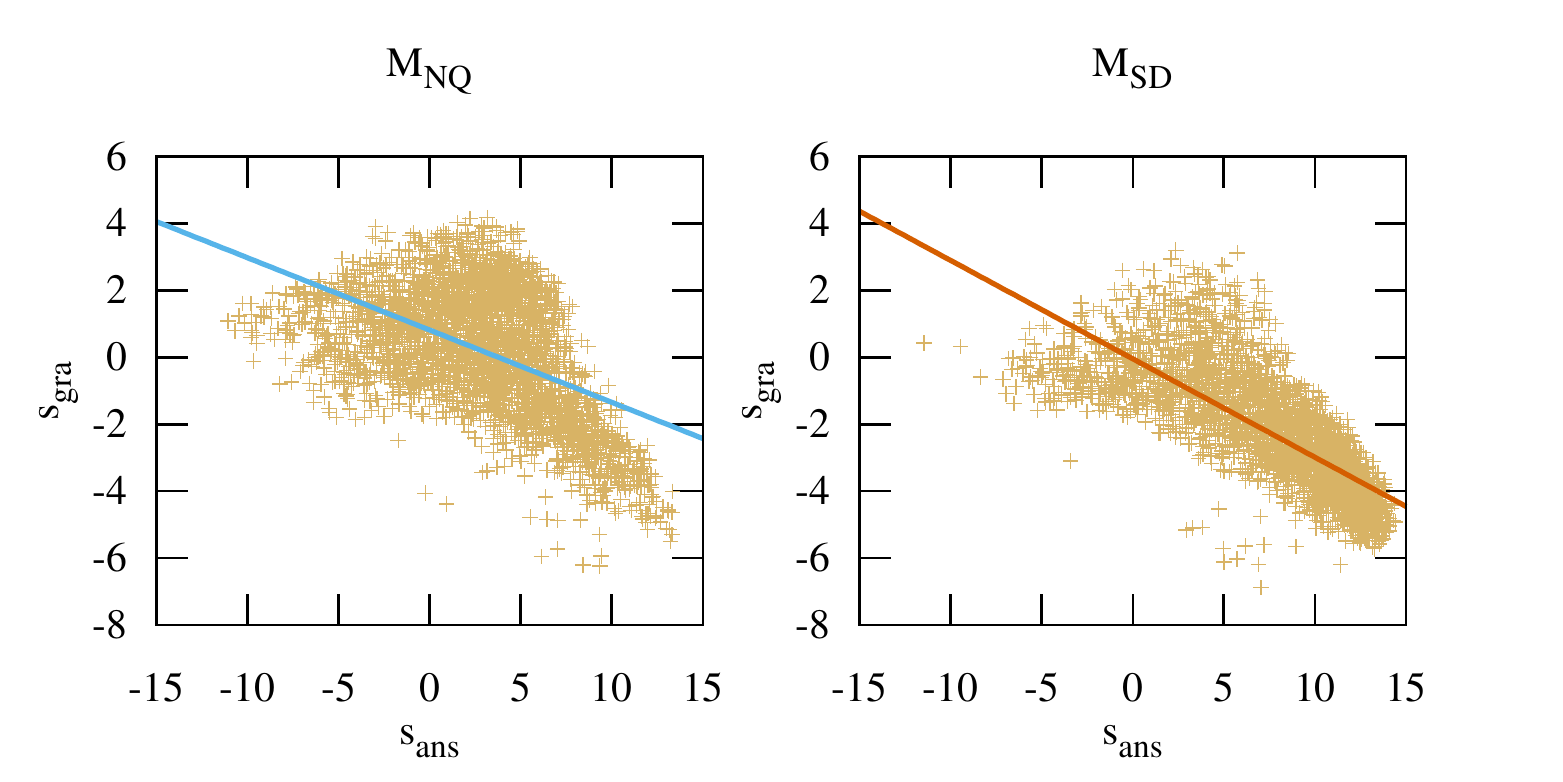}
\caption{Scatter plots of generated questions of the News-LA from $M_{NQ}$ (left) and $M_{SD}$ (right). $s_{ans}$ and $s_{gra}$ are negatively correlated, but the $M_{NQ}$ model tends to generate more questions with positive anserability and granularity. Straight lines show fitted linear regressions.}
\label{fig:scatter-ans-vs-gra}
\end{figure}

We show the QA evaluation results in Figure \ref{fig:newsqa-scores}. In the context column, $M_{NQ}$ shows a lower answerability score than the baseline model $M_{SD}$. 
While granularity scores show a reverse trend, i.e., higher scores for $M_{NQ}$ than those of $M_{SD}$.
This result implies that $M_{NQ}$ generates more coarse-style questions that have long answers, but these questions are considerably more difficult to answer by the QA model, comparing to short-answer questions.

It is also reasonable to assume that news articles' summaries are proper answer-candidates for long-answer questions. 
Highlights in news articles are human-generated summaries, so we also combine the same set of questions with their corresponding highlights as input for the \Bert-joint QA system with results shown as the highlights column in Figure \ref{fig:newsqa-scores}. 
The answerability scores drop for both models comparing the column highlights to the column of context, which is reasonable as the models never see highlights when generating questions.
However, the baseline method $M_{SD}$ drops more significantly than $M_{NQ}$, suggesting that the baseline model is more context-dependent while our model $M_{NQ}$ generates more self-explanatory questions.
From the granularity scores of highlights, we find that confidence to determine answer types is lower for both models than that of the context column. However, the $M_{NQ}$ still shows higher granularity scores than the $M_{SD}$.

We map generated questions for the News-LA on a 2D plot with x-axis the answerability score and y-axis the granularity score for both models in Figure \ref{fig:scatter-ans-vs-gra}. They also confirm the negative correlation between answerability and granularity of generated questions. However, the $M_{NQ}$ generates more questions with both positive $s_{ans}$ and $s_{gra}$ than those from $M_{SD}$, indicating the effectiveness of our model to generate introductory and self-explanatory questions.

\section{Out-of-Domain Human Evaluation}
\label{sec:human-eval}

\begin{table}[ht]
\centering
\small
\begin{tabular}{lS[table-format=2]S[table-format=2]S[table-format=2]S[table-format=2]S[table-format=2]S[table-format=2]}
\toprule
 (\si{\percent})  & \multicolumn{2}{r}{Context} & \multicolumn{2}{r}{Span} & \multicolumn{2}{r}{Entire} \\
   & {T}             & {F}            & {T}            & {F}          & {T}            & {F}            \\
\midrule
$M_{NQ}$ & 38            & 62          & 77          & 23         & 49          & 51          \\
$M_{SD}$ & 70           & 30           & 89          & 11         & 40           & 60          \\
\bottomrule
\end{tabular}
\caption{Ratios (shown as a percentage) between {\it True} and {\it False} for human evaluation with three statements ({\it Context}, {\it Span}, and {\it Entire}) on generated questions. We count true/false marked by annotators with unanimity amongst all three annotators for each statement.}
\label{tbl:human-eval-three-factors}
\end{table}

\begin{table}[t]
\centering
\small
\begin{tabular}{lS[table-format=-2.1]S[table-format=-2.1]S[table-format=-2.1]S[table-format=-2.1]}
\toprule
 & \multicolumn{2}{c}{$s_{ans}$} & \multicolumn{2}{c}{$s_{gra}$} \\
& $M_{NQ}$ & $M_{SD}$ & $M_{NQ}$ & $M_{SD}$ \\
\midrule
Context       &  0.1  & -0.1  &  0.1  &  0.5 \\ 
Irrelevant    & -1.0  & -0.6  &  \bfseries 0.7  &  0.4 \\ 
Contradiction & -0.5  & -0.3  &  0.4  &  0.2 \\ 
Peripheral    & -0.3  & -0.3  &  0.2  &  0.2 \\ 
Span          &  \bfseries 1.5  &  \bfseries 1.1  & \bfseries -0.8  & \bfseries -0.6 \\ 
Entire        &  0.4  &  0.3  &  0.4  &  0.3 \\ 
None          & \bfseries -1.5  & \bfseries -1.2  &  0.6  &  \bfseries 0.6 \\ 
\bottomrule
\end{tabular}
\caption{Pearson correlation (\num{1e-1}) between human (Section \ref{sec:human-eval}) and automatic (Section \ref{sec:newsqa-la-evaluation}) evaluation. For each column, we mark the most positive and negative correlated scores in bold text.}
\label{tbl:correlation-human-auto}
\end{table}

We further conduct a human evaluation using MTurk for the News-LA test set to verify that we can generate self-explanatory and introductory questions and that the automatic evaluation in Section \ref{sec:newsqa-la-evaluation} agrees with human evaluation. We ask annotators to read news articles and mark true or false for seven statements regarding generated questions. For each context-question pair, these statements include (see examples in Appendix \ref{sec:app:human-eval})
\begin{itemize}
    \item Question is {\it context} dependent
    \item Question is {\it irrelevant} to the article
    \item Question implies a {\it contradiction} to facts present in the article
    \item Question focuses on a {\it peripheral} topic
    \item There is a short {\it span} to answer the question
    \item The {\it entire} article can be an answer
    \item {\it None} answer in the article
\end{itemize}

We randomly select \num{1000} news articles in News-LA to perform our human evaluation with three different annotators per news article. We received three valid annotations for \num{943} news articles from a set of \num{224} annotators. We first consider true/false results regarding three metrics -- {\it Context}, {\it Span}, and {\it Entire} -- considering only when unanimity is reached among annotators (Table \ref{tbl:human-eval-three-factors}). 
$M_{NQ}$ questions are more context-free than $M_{SD}$ ones, with \SI{38}{\percent} true and \SI{62}{\percent} false towards the {\it Context} statement.
Second, the $M_{NQ}$ questions are more likely to be answered by entire news articles (\SI{49}{\percent} true of {\it Entire} vs. \SI{40}{\percent}) while less likely to be answered by spans from news articles (\SI{77}{\percent} true of {\it Span} vs. \SI{89}{\percent}) comparing with $M_{SD}$ questions. 
These human evaluation results confirm that $M_{NQ}$ questions are more self-explanatory and introductory than $M_{SD}$.

We compute the $s_{ans}$ and $s_{gra}$ for the \num{943} generated questions (Section \ref{sec:newsqa-la-evaluation}). We then normalize these two scores and conduct a Pearson correlation analysis \cite{benesty2009pearson} with human evaluation results.
We use all human evaluation results, regardless of agreements among annotators.  
From Table \ref{tbl:correlation-human-auto}, we find that {\it Span} has the strongest positive correlation with the $s_{ans}$, while {\it None} shows the strongest negative correlation -- aligning with the findings for answerability. {\it Span} also shows the strongest negative correlation with the $s_{gra}$ for both $M_{NQ}$ and $M_{SD}$, but the highest positive correlation with granularity varies, with {\it Irrelevant} for $M_{NQ}$ questions and {\it None} for $M_{SD}$ questions.

\section{Conclusion}

We tackle the problem of question generation targeted for human information seeking using automatic question answering technology. We focus on generating questions for news articles that can be answered by longer passages rather than short text spans as suggested questions. We build a \Bert-based Pointer-Generator Network as the QG model, trained with the Natural Questions dataset. Our method shows state-of-the-art performance in terms of BLEU, METEOR, and ROUGE\_L scores on our NQ question generation dataset. 
We then apply our model to the out-of-domain news articles without further training. 
We use a QA system to evaluate our QG models as there are no gold questions for comparison. 
We also conduct a human evaluation to confirm the QA evaluation results.

\section*{Broader Impact}
We describe a method for an autonomous agent to suggest questions based on machine-reading and question generation technology. Operationally, this work focuses on newswire-sourced data where the generated questions are answered by the text -- and is applicable to multi-turn search settings. Thus, there are several potentially positive social impacts. By presenting questions with known answers in the text, users can more efficiently learn about topics in the source documents. Our focus on {\em self-explanatory} and {\em introductory} questions increases the utility of questions for this purpose. 

Conversely, there is potential to bias people toward a subset of the news chosen by a purported fair search engine, which may be more difficult to detect as the provided questions remove some of the article contexts. In principle, this is mitigated by selecting content that maintains high journalistic standards -- but such a risk remains if the technology is deployed by bad-faith actors.

The data for our experiments was derived from the widely used Natural Questions~\cite{kwiatkowshi-etal-2019-natural-questions} and CNN/Daily Mail~\cite{10.5555/2969239.2969428} datasets, which in turn were derived from public news sourced data. Our evaluation annotations were performed on Amazon Mechanical Turk, where three authors completed a sample task and set a wage corresponding to an expected rate of \SI[per-mode=symbol]{15}{\$\per\hour}.
\bibliography{anthology,aaai21}
\bibliographystyle{acl_natbib}

%
%
\cleardoublepage
\newpage
\appendix

\section{Appendix}
\label{sec:appendix}

\begin{figure*}[ht]
    \centering
    \small
    \begin{tabular}{|l|}
    \hline
\begin{minipage}[t]{0.85\textwidth}%
\textcolor{orange}{President of the United Nations General Assembly}  \textcolor{cyan}{[ Miroslav Lajčák of Slovakia ]} has been elected as the United Nations General Assembly President of its 72nd session beginning in September 2017.
\end{minipage} \\\hline

\textit{who is the current president of un general assembly} \\\hline\hline

\begin{minipage}[t]{0.85\textwidth}%
\textcolor{orange}{Learner 's permit}  Typically , a driver operating with a learner 's permit must be accompanied by \textcolor{cyan}{[ an adult licensed driver who is at least 21 years of age or older and in the passenger seat of the vehicle at all times ]} .
\end{minipage} \\\hline

\textit{who needs to be in the car with a permit driver} \\\hline\hline

\begin{minipage}[t]{0.85\textwidth}%
\textcolor{orange}{Java development Kit} \textcolor{cyan}{[  The Java Development Kit ( JDK ) is an implementation of either one of the Java Platform , Standard Edition , Java Platform , Enterprise Edition , or Java Platform , Micro Edition platforms released by Oracle Corporation in the form of a binary product aimed at Java developers on Solaris , Linux , macOS or Windows . The JDK includes a private JVM and a few other resources to finish the development of a Java Application . Since the introduction of the Java platform , it has been by far the most widely used Software Development Kit ( SDK ) . On 17 November 2006 , Sun announced that they would release it under the GNU General Public License ( GPL ) , thus making it free software . This happened in large part on 8 May 2007 , when Sun contributed the source code to the OpenJDK .  ]}

\end{minipage} \\\hline

\textit{what is the use of jdk in java} \\\hline

\end{tabular}
    \caption{Examples of the NQ data preprocessing from the training set. Orange texts are Wikipedia titles that added in the front the each long answers. In first two examples, annotators mark there are short answers represented in cyan; while for the last example, there is no short answer marked by annotators so we mark the whole paragraph as the answer. Cyan texts are tagged with type ID `\num{1}' during preprocessing.}
    \label{fig:example-of-preprocessing}
\end{figure*}

\begin{figure*}[ht]
    \centering
    \small
    \begin{tabular}{|l|l|}
    \hline
    Context &
\begin{minipage}[t]{0.85\textwidth}%
Therefore sign \textbf{[~(1)} In logical argument and mathematical proof, \textbf{[~(2)} \textbf{[~(3)} the \textbf{[~(4)} therefore sign \textbf{(/4)~]} \textbf{(/3)~]} ( $\therefore$ ) is generally used before \textbf{[~(5)} a logical consequence, such as the conclusion of a syllogism. \textbf{(/5)~]} \textbf{(/2)~]} The symbol consists of three dots placed in an upright triangle and is read therefore. It is encoded at U+2234 $\therefore$ therefore (HTML \&\#8756; \&there4;). For common use in Microsoft Office hold the ALT key and type ``8756''. While it is not generally used in formal writing, it is used in mathematics and shorthand. It is complementary to U+2235 $\because$ because (HTML \&\#8757;). \textbf{(/1)~]}
\end{minipage} \\

Question & \textit{what do the 3 dots mean in math} \\\hline\hline

SA 1 &  whole paragraph \\
Predicted & \textit{what is the therefore sign in a syllogism} \\\hline
SA 2 & 
\begin{minipage}[t]{0.85\textwidth}%
[the therefore sign ( $\therefore$ ) is generally used before a logical consequence, such as the conclusion of a syllogism.]
\end{minipage} \\
Predicted & \textit{what is the meaning of therefore in triangle}\\\hline
SA 3 &  [the therefore sign] \\
Predicted & \textit{what is the name of the three dots in a triangle called} \\\hline
SA 4 & [therefore sign] \\
Predicted & \textit{what is the name of the three dots in a triangle called} \\\hline
SA 5 & [a logical consequence , such as the conclusion of a syllogism] \\
Predicted & \textit{when is the therefore sign used in a syllogism} \\\hline
\end{tabular}
    \caption{Example of the question generation from Natural Questions dataset with \BertPGN. We use `[~(i)' and `(/i)~]' to represent the start and end position of the \textit{i}-th answer span. The context is the long answer for the question {\it what do the 3 dots mean in math}. Five short answers (SA) marked by five different annotators. Our \BertPGN model with nucleus sampling \cite{holtzman2019curious} with temperature of \num{0.1} produces different but related questions for each short answers as well as the whole context with brackets over each of them.}
    \label{fig:bert-pgn-example}
\end{figure*}

We show several generated questions here. Each frame box contains a news article, with two questions generated by $M_{NQ}$ (showing in bold texts) and $M_{SD}$ respectively.
News articles are selected from the CNN/Daily Mail dataset with preprocessing described in Section \ref{sec:newsqa-data}.
We also compare these generated questions in Table \ref{tbl:generated-questions}.

\begin{table*}[t]
\centering
\small
\begin{tabular}{p{0.45\linewidth} | p{0.45\linewidth}}
\toprule
\BertPGN-NQ-whole-article & \BertPGN-SQuAD-first-line  \\
\midrule
who are the new astronauts on the moon                    &  how many italians walk into a space station in 2013  \\
when is the space shuttle discovery coming out            &  how many days is the space shuttle discovery scheduled to launch  \\
what is the average unemployment rate in spain            &  what percentage of spain's population is out of work  \\
what is the meaning of soulja boy tell em                 &  what was deandre cortez way known as
 \\
where does the us refugees at guantanamo bay come from    &  what is the name of the us military facility in the us
 \\
what happened to the girl in the texas polygamist ranch   & what was the name of the texas polygamist ranch
  \\
who scored the first goal in the premier league           &  which team did everton fc beat to win the premier league's home draw with tottenham on sunday
 \\

\bottomrule
\end{tabular}
\caption{Comparing generated questions side-by-side. Our model uses uncased vocabulary and omits the final question mark.}
\label{tbl:generated-questions}
\end{table*}

\begin{mdframed}
\small
Two Italians, a Dane, a German, a Frenchman and a Brit walk into a space station... or will, in 2013, if all goes according to European Space Agency plans. Europe's six new astronauts hope to join their American counterparts on the Internation Space Station. The six new astronauts named Wednesday were chosen from more than 8,400 candidates, and are the first new ESA astronauts since 1992, the space agency said in a statement. They include two military test pilots, one fighter pilot and one commercial pilot, plus an engineer and a physicist. "This is a very important day for human spaceflight in Europe," said Simonetta Di Pippo, Director of Human Spaceflight at ESA. "These young men and women are the next generation of European space explorers. They have a fantastic career ahead, which will put them right on top of one of the ultimate challenges of our time: going back to the Moon and beyond as part of the global exploration effort." Humans have not walked on the moon since 1972, just over three years after the first manned mission to Earth's nearest neighbor. The six will begin space training in Germany, with an eye to being ready for future missions to the International Space Station and beyond in four years. They are: Samantha Cristoforetti of Italy, a fighter pilot with degrees in engineering and aeronautical sciences; Alexander Gerst, a German researcher with degrees in physics and earth science; Andreas Mogensen, a Danish engineer with the private space firm HE Space Operations; Luca Parmitano of Italy, an Air Force pilot with a degree in aeronautical sciences; Timothy Peake, an English test pilot with the British military; and Frenchman Thomas Pesquet, an Air France pilot who previously worked as an engineer at the French space agency.
\end{mdframed}

\begin{itemize}
    \item \textbf{who are the new astronauts on the moon}
    \item how many italians walk into a space station in 2013
\end{itemize}

\begin{mdframed}
\small
After several delays, NASA said Friday that space shuttle Discovery is scheduled for launch in five days. The space shuttle Discovery, seen here in January, is now scheduled to launch Wednesday. Commander Lee Archambault and his six crewmates are now scheduled to lift off to the International Space Station at 9:20 p.m. ET Wednesday. NASA said its managers had completed a readiness review for Discovery, which will be making the 28th shuttle mission to the ISS. The launch date had been delayed to allow "additional analysis and particle impact testing associated with a flow-control valve in the shuttle's main engines," the agency said. According to NASA, the readiness review was initiated after damage was found in a valve on the shuttle Endeavour during its November 2008 flight. Three valves have been cleared and installed on Discovery, it said. Discovery is to deliver the fourth and final set of "solar array wings" to the ISS. With the completed array the station will be able to provide enough electricity when the crew size is doubled to six in May, NASA said. The Discovery also will carry a replacement for a failed unit in a system that converts urine to drinkable water, it said. Discovery's 14-day mission will include four spacewalks, NASA said.
\end{mdframed}

\begin{itemize}
    \item \textbf{when is the space shuttle discovery coming out}
    \item how many days is the space shuttle discovery scheduled to launch
\end{itemize}

\begin{mdframed}
\small
Unemployment in Spain has reached 20 percent, meaning 4.6 million people are out of work, the Spanish government announced Friday. The figure, from the first quarter, is up from 19 percent and 4.3 million people in the previous quarter. It represents the second-highest unemployment rate in the European Union, after Latvia, according to figures Friday from Eurostat, the EU's statistics service. Spanish Prime Minister Jose Luis Rodriguez Zapatero told Parliament on Wednesday he believes the jobless rate has peaked and will now start to decline. The first quarter of the year is traditionally poor for Spain because of a drop in labor-intensive activity like construction, agriculture and tourism. This week, Standard \& Poor's downgraded Spain's long-term credit rating and said the outlook is negative. "We now believe that the Spanish economy's shift away from credit-fuelled economic growth is likely to result in a more protracted period of sluggish activity than we previously assumed," Standard \& Poor's credit analyst Marko Mrsnik said. Gross domestic product growth in Spain is expected to average 0.7 percent annually through 2016, compared with previous expectations of 1 percent annually, he said. Spain's economic problems are closely tied to the housing bust there, according to The Economist magazine. Many of the newly unemployed worked in construction, it said. The recession revealed how dependent public finances were on housing-related tax revenues, it said. Another problem in Spain is that wages are set centrally and most jobs are protected, making it hard to shift skilled workers from one industry to another, the magazine said. Average unemployment for the 27-member European Union stayed stable in March at 9.6 percent, Eurostat said Friday. That percentage represents 23 million people, it said. The lowest national unemployment rates were in the Netherlands and Austria, which had 4.1 and 4.9 percent respectively, Eurostat said.
\end{mdframed}

\begin{itemize}
    \item \textbf{what is the average unemployment rate in spain}
    \item what percentage of spain's population is out of work 
\end{itemize}
    
\begin{mdframed}
\small
Atlanta rapper DeAndre Cortez Way, better known by his stage name Soulja Boy Tell 'Em or just Soulja Boy, was charged with obstruction after running from police despite an order to stop, a police spokesman said Friday. Rapper Soulja Boy was arrested in Georgia after allegedly running from police. The 19-year-old singer was among a large group that had gathered at a home in Stockbridge, 20 miles south of Atlanta, said Henry County, Georgia, police Capt. Jason Bolton. Way was arrested Wednesday night along with another man, Bolton said. Police said Way left jail Thursday after posting a \$550 bond. Bolton said officers responded to a complaint about a group of youths milling around the house, which appeared to be abandoned. When police arrived, they saw about 40 people. Half of them ran away, including Way, Bolton said. The ones who remained told officers they were at the home to film a video. Way was arrested when he returned to the house to get his car, Bolton said. He said the house was dark inside and looked abandoned. "He just ran from the police, and then he decided to come back," according to Bolton. The second man who returned for his vehicle was arrested after police found eight \$100 counterfeit bills inside, according to the officer. Way broke into the music scene two years ago with his hit "Crank That (Soulja Boy)." The rapper also describes himself as a producer and entrepreneur.
\end{mdframed}

\begin{itemize}
    \item \textbf{what is the meaning of soulja boy tell em}
    \item what was deandre cortez way known as
\end{itemize}
      
\begin{mdframed}
\small
The U.S. military is gearing up for a possible influx of Haitians fleeing their earthquake-stricken country at an Army facility not widely known for its humanitarian missions: Guantanamo Bay. Soldiers at the base have set up tents, beds and toilets, awaiting possible orders from the secretary of defense to proceed, according to Maj. Diana Haynie, a spokeswoman for Joint Task Force Guantanamo Bay. "There's no indication of any mass migration from Haiti," Haynie stressed. "We have not been told to conduct migrant operations." But the base is getting ready "as a prudent measure," Haynie said, since "it takes some time to set things up." Guantanamo Bay is about 200 miles from Haiti. Currently, military personnel at the base are helping the earthquake relief effort by shipping bottled water and food from its warehouse. In addition, Gen. Douglas Fraser, commander of U.S. Southern Command, said the Navy has set up a "logistics field," an area to support bigger ships in the region. The military can now use that as a "lily pad" to fly supplies from ships docked at Guantanamo over to Haiti, he said. "Guantanamo Bay proves its value as a strategic hub for the movement of supplies and personnel to the affected areas in Haiti," Haynie said. As part of the precautionary measures to prepare for possible refugees, the Army has erected 100 tents, each holding 10 beds, according to Haynie. Toilet facilities are nearby. If needed, hundreds more tents are stored in Guantanamo Bay and can be erected, she said. The refugees would be put on the leeward side of the island, more than 2 miles from some 200 detainees being held on the other side, Haynie said. The refugees would not mix with the detainees. Joint Task Force Guantanamo Bay is responsible for planning for any kind of Caribbean mass immigration, according to Haynie. In the early 1990s, thousands of Haitian refugees took shelter on the island, she said.
\end{mdframed}

\begin{itemize}
    \item \textbf{where does the us refugees at guantanamo bay come from}
    \item what is the name of the us military facility in the us
\end{itemize}

\begin{mdframed}
\small
A Colorado woman is being pursued as a "person of interest" in connection with phone calls that triggered the raid of a Texas polygamist ranch, authorities said Friday. Rozita Swinton, 33, has been arrested in a case that is not directly related to the Texas raid. Texas Rangers are seeking Rozita Swinton of Colorado Springs, Colorado, "regarding telephone calls placed to a crisis center hot line in San Angelo, Texas, in late March 2008," the Rangers said in a written statement. The raid of the YFZ (Yearning for Zion) Ranch in Eldorado, Texas, came after a caller -- who identified herself as a 16-year-old girl -- said she had been physically and sexually abused by an adult man with whom she was forced into a "spiritual marriage." The release said a search of Swinton's home in Colorado uncovered evidence that possibly links her to phone calls made about the ranch, run by the Fundamentalist Church of Jesus Christ of Latter-day Saints. "The possibility exists that Rozita Swinton, who has nothing to do with the FLDS church, may have been a woman who made calls and pretended she was the 16-year-old girl named Sarah," CNN's Gary Tuchman reported. Swinton, 33, has been charged in Colorado with false reporting to authorities and is in police custody. Police said that arrest was not directly related to the Texas case. Authorities raided the Texas ranch April 4 and removed 416 children. Officials have been trying to identify the 16-year-old girl, referred to as Sarah, who claimed she had been abused in the phone calls. FLDS members have denied the girl, supposedly named Sarah Jessop Barlow, exists. Some of the FLDS women who spoke with CNN on Monday said they believed the calls were a hoax. While the phone calls initially prompted the raid, officers received a second search warrant based on what they said was evidence of sexual abuse found at the compound. In court documents, investigators described seeing teen girls who appeared pregnant, records that showed men marrying multiple women and accounts of girls being married to adult men when they were as young as 13. A court hearing began Thursday to determine custody of children who were removed from the ranch.
\end{mdframed}

\begin{itemize}
    \item \textbf{what happened to the girl in the texas polygamist ranch}
    \item what was the name of the texas polygamist ranch
\end{itemize}

\begin{mdframed}
\small
Everton scored twice late on and goalkeeper Tim Howard saved an injury-time penalty as they fought back to secure a 2-2 Premier League home draw with Tottenham on Sunday. Jermain Defoe gave the visitors the lead soon after the interval when nipping in front of Tony Hibbert to convert Aaron Lennon's cross at the near post for his 13th goal of the season. And they doubled their advantage soon after when defender Michael Dawson headed home a Niko Kranjcar corner. But Everton got a foothold back in the game when Seamus Coleman's run and cross was converted by fellow-substitute Louis Saha in the 78th minute. And Tim Cahill rescued a point for the home side with four minutes remaining when he stooped low to head home Leighton Baines' bouncing cross. However, there was still further drama to come when Hibbert was penalized for crashing into Wilson Palacios in the area. However, England striker Defoe smashed his penalty too close to Howard and the keeper pulled off a fine save to give out-of-form Everton a morale-boosting point. The result means Tottenham remain in fourth place, behind north London rivals Arsenal, while Everton have now won just one of their last nine league games. In the day's other match, Bobby Zamora scored the only goal of the game as Fulham beat Sunderland 1-0 to move up to eighth place in the table.
\end{mdframed}

\begin{itemize}
    \item \textbf{who scored the first goal in the premier league}
    \item which team did everton fc beat to win the premier league's home draw with tottenham on sunday
\end{itemize}

\section{Human Evaluation Criteria}
\label{sec:app:human-eval}

\noindent\textbf{Question is context dependent}

Some questions are context-dependent, e.g.,

\begin{itemize}
    \item “who intends to boycott the election” - which election?
    \item “where did the hijackers go to” - what hijackers?
    \item “what type of hats did they use” - who are they?
    \item “how many people were killed in the quake” - which quake?
\end{itemize}

Compared to these context-independent, self-contained questions:

\begin{itemize}
    \item “what was toyota's first-ever net loss”
    \item “who is hillary's secretary of state”
    \item “what is the name of the motto of the new york times ”
\end{itemize}

\bigskip
\noindent\textbf{Question is irrelevant to the article}

Given a news article:

\begin{mdframed}
\small
"Usually when I mention suspended animation people will flash me the Vulcan sign and laugh," says scientist Mark Roth. But he's not referring to the plot of a "Star Trek" episode. Roth is completely serious about using lessons he's learned from putting some organisms into suspended animation to help people survive medical trauma. He spoke at the TED2010 conference in Long Beach, California, in February. The winner of a MacArthur genius fellowship in 2007, Roth described the thought process that led him and fellow researchers to explore ways to lower animals' metabolism to the point where they showed no signs of life -- and yet were not dead. More remarkably, they were able to restore the animals to normal life, with no apparent damage. Read more about Roth on TED.com The Web site of Roth's laboratory at the Fred Hutchinson Cancer Research Center in Seattle, Washington, describes the research this way: "We use the term suspended animation to refer to a state where all observable life processes (using high resolution light microscopy) are stopped: The animals do not move nor breathe and the heart does not beat. We have found that we are able to put a number of animals (yeast, nematodes, drosophila, frogs and zebrafish) into a state of suspended animation for up to 24 hours through one basic technique: reducing the concentration of oxygen." Visit Mark Roth's laboratory Roth is investigating the use of small amounts of hydrogen sulfide, a gas that is toxic in larger quantities, to lower metabolism. In his talk, he imagined that "in the not too distant future, an EMT might give an injection of hydrogen sulfide, or some related compound, to a person suffering severe injuries, and that person might de-animate a bit ... their metabolism will fall as though you were dimming a switch on a lamp at home. "That will buy them the time to be transported to the hospital to get the care they need. And then, after they get that care ... they'll wake up. A miracle? We hope not, or maybe we just hope to make miracles a little more common."
\end{mdframed}

The question: “what is the meaning of suspended animation in star trek” is irrelevant to the news since the news is not talking about Star Trek.

However, the question “what is the meaning of suspended animation” is related.

\bigskip
\noindent\textbf{Question implies a contradiction to facts present in the article}

Given a news article:

\begin{mdframed}
\small
At least 6,000 Christians have fled the northern Iraqi city of Mosul in the past week because of killings and death threats, Iraq's Ministry of Immigration and Displaced Persons said Thursday. A Christian family that fled Mosul found refuge in the Al-Sayida monastery about 30 miles north of the city. The number represents 1,424 families, at least 70 more families than were reported to be displaced on Wednesday. The ministry said it had set up an operation room to follow up sending urgent aid to the displaced Christian families as a result of attacks by what it called "terrorist groups." Iraqi officials have said the families were frightened by a series of killings and threats by Muslim extremists ordering them to convert to Islam or face death. Fourteen Christians have been slain in the past two weeks in the city, which is about 260 miles (420 kilometers) north of Baghdad. Mosul is one of the last Iraqi cities where al Qaeda in Iraq has a significant presence and routinely carries out attacks. The U.S. military said it killed the Sunni militant group's No. 2 leader, Abu Qaswarah, in a raid in the northern city earlier this month. In response to the recent attacks on Christians, authorities have ordered more checkpoints in several of the city's Christian neighborhoods. The attacks may have been prompted by Christian demonstrations ahead of provincial elections, which are to be held by January 31, authorities said. Hundreds of Christians took to the streets in Mosul and surrounding villages and towns, demanding adequate representation on provincial councils, whose members will be chosen in the local elections. Thursday, Iraq's minister of immigration and displaced persons discussed building housing complexes for Christian families in northern Iraq and allocating land to build the complexes. Abdel Samad Rahman Sultan brought up the issue when he met with a representative of Iraq's Hammurabi Organization for Human Rights and with the head of the Kojina Organization for helping displaced persons. A curfew was declared Wednesday in several neighborhoods of eastern Mosul as authorities searched for militants behind the attacks.
\end{mdframed}

The question “how many christians fled to mosul in the past” is contradicted to the fact --- \num{6000} christians fled from Mosul --- in the news.

\bigskip
\noindent\textbf{Question focuses on a peripheral topic}

Given a news article: 

\begin{mdframed}
\small
One of the Marines shown in a famous World War II photograph raising the U.S. flag on Iwo Jima was posthumously awarded a certificate of U.S. citizenship on Tuesday. The Marine Corps War Memorial in Virginia depicts Strank and five others raising a flag on Iwo Jima. Sgt. Michael Strank, who was born in Czechoslovakia and came to the United States when he was 3, derived U.S. citizenship when his father was naturalized in 1935. However, U.S. Citizenship and Immigration Services recently discovered that Strank never was given citizenship papers. At a ceremony Tuesday at the Marine Corps Memorial -- which depicts the flag-raising -- in Arlington, Virginia, a certificate of citizenship was presented to Strank's younger sister, Mary Pero. Strank and five other men became national icons when an Associated Press photographer captured the image of them planting an American flag on top of Mount Suribachi on February 23, 1945. Strank was killed in action on the island on March 1, 1945, less than a month before the battle between Japanese and U.S. forces there ended. Jonathan Scharfen, the acting director of CIS, presented the citizenship certificate Tuesday. He hailed Strank as "a true American hero and a wonderful example of the remarkable contribution and sacrifices that immigrants have made to our great republic throughout its history."
\end{mdframed}

The question “who presented the american flag raising on iwo jima” focuses on a peripheral topic --- the name of the one raising the flag.

While the question “who was awarded a certificate of citizenship raising the u.s. flag” focuses on the main topic - getting a citizenship.

\bigskip
\noindent\textbf{There is a short span to answer the question}

Given a news:

\begin{mdframed}
\small
Los Angeles police have launched an internal investigation to determine who leaked a picture that appears to show a bruised and battered Rihanna. Rihanna was allegedly attacked by her boyfriend, singer Chris Brown, before the Grammys on February 8. The close-up photo -- showing a woman with contusions on her forehead and below her eyes, and cuts on her lip -- was published on the entertainment Web site TMZ Thursday. TMZ said it was a photo of Rihanna. Twenty-one-year-old Rihanna was allegedly attacked by her boyfriend, singer Chris Brown, on a Los Angeles street before the two were to perform at the Grammys on February 8. "The unauthorized release of a domestic violence photograph immediately generated an internal investigation," an L.A. police spokesman said in a statement. "The Los Angeles Police Department takes seriously its duty to maintain the confidentiality of victims of domestic violence. A violation of this type is considered serious misconduct, with penalties up to and including termination." A spokeswoman for Rihanna declined to comment. The chief investigator in the case had told CNN earlier that authorities had tried to guard against leaks. Detective Deshon Andrews said he had kept the case file closely guarded and that no copies had been made of the original photos and documents. Brown was arrested on February 8 in connection with the case and and booked on suspicion of making criminal threats. Authorities are trying to determine whether Brown should face domestic violence-related charges. Brown apologized for the incident this week. "Words cannot begin to express how sorry and saddened I am over what transpired," the 19-year-old said in a statement released by his spokesman. "I am seeking the counseling of my pastor, my mother and other loved ones and I am committed, with God's help, to emerging a better person."
\end{mdframed}

The question “who have launched an internal investigation of the leaked rihanna's picture” can be answered by “Los Angeles police”.

\bigskip
\noindent\textbf{The entire article can be an answer}

Given a news:

\begin{mdframed}
\small
A high court in northern India on Friday acquitted a wealthy businessman facing the death sentence for the killing of a teen in a case dubbed "the house of horrors." Moninder Singh Pandher was sentenced to death by a lower court in February. The teen was one of 19 victims -- children and young women -- in one of the most gruesome serial killings in India in recent years. The Allahabad high court has acquitted Moninder Singh Pandher, his lawyer Sikandar B. Kochar told CNN. Pandher and his domestic employee Surinder Koli were sentenced to death in February by a lower court for the rape and murder of the 14-year-old. The high court upheld Koli's death sentence, Kochar said. The two were arrested two years ago after body parts packed in plastic bags were found near their home in Noida, a New Delhi suburb. Their home was later dubbed a "house of horrors" by the Indian media. Pandher was not named a main suspect by investigators initially, but was summoned as co-accused during the trial, Kochar said. Kochar said his client was in Australia when the teen was raped and killed. Pandher faces trial in the remaining 18 killings and could remain in custody, the attorney said.
\end{mdframed}

The question “what was the case of the house of horrors in northern india” can be answered by the whole news article. There is no short span can be extracted as an answer.

\bigskip
\noindent\textbf{None answer in the article}

Given a news:

\begin{mdframed}
\small
Buy a \$175,000 package to attend the Oscars and you might buy yourself trouble, lawyers for the Academy Awards warn. The 81st annual Academy Awards will be held on February 22 from Hollywood's Kodak Theatre. The advertising of such packages -- including four tickets to the upcoming 81st annual Academy Awards and a hotel stay in Los Angeles, California -- has prompted the Academy of Motion Picture Arts and Sciences to sue an Arizona-based company. The Academy accused the company Experience 6 of selling "black-market" tickets, because tickets to the lavish movie awards show cannot be transferred or sold. Selling tickets could become a security issue that could bring celebrity stalkers or terrorists to the star-studded event, says the lawsuit, which was filed Monday in federal court in the Central District of California. "Security experts have advised the Academy that it must not offer tickets to members of the public and must know identities of the event attendees," the lawsuit says. "In offering such black-market tickets, defendants are misleading the public and the ticket buyers into thinking that purchasers will be welcomed guests, rather than as trespassers, when they arrive for the ceremony." Experience 6 did not return calls from CNN for comment. On Tuesday morning, tickets to the event were still being advertised on the company's Web site. The Oscars will be presented February 22 from Hollywood's Kodak Theatre. The Academy Awards broadcast will air on ABC. Hugh Jackman is scheduled to host.
\end{mdframed}

The questions “where does the \num{81}st annual academy awards come from” and “how much did the academy pay to attend the oscars” cannot be answered from the news.

\end{document}